\documentclass{article}
\usepackage{multirow} 
\usepackage{microtype}
\usepackage{graphicx}
\usepackage{subfigure}
\usepackage{booktabs} 

\usepackage{hyperref}

\usepackage[accepted]{icml2024}

\usepackage{amsmath}
\usepackage{amssymb}
\usepackage{mathtools}
\usepackage{amsthm}

\usepackage[capitalize,noabbrev]{cleveref}

\theoremstyle{plain}

\theoremstyle{definition}

\theoremstyle{remark}

\usepackage[textsize=tiny]{todonotes}

\icmltitlerunning{Sparse Inducing Points in DGPs: Enhancing Modeling with DDVI}

\begin{document}
	
	\twocolumn[
	\icmltitle{Sparse Inducing Points in Deep Gaussian Processes:
		\\
		Enhancing Modeling with Denoising Diffusion Variational Inference}
	
	
	

	\begin{icmlauthorlist}
		\icmlauthor{Jian Xu}{yyy}
		\icmlauthor{Delu Zeng}{yyy}
		\icmlauthor{John Paisley}{comp}
		
	\end{icmlauthorlist}
	
	\icmlaffiliation{yyy}{South China University of Technology, Guangzhou, China}
	
	\icmlaffiliation{comp}{Columbia University, New York, USA}

	\icmlcorrespondingauthor{Delu Zeng}{dlzeng@scut.edu.cn}

	\icmlkeywords{Gaussian processes, denoising diffusions, variational inference}
	
	\vskip 0.3in
	]
	
	
	
	\printAffiliationsAndNotice{}  
	
	\begin{abstract}
		Deep Gaussian processes (DGPs) provide a robust paradigm for Bayesian deep learning. In DGPs, a set of sparse integration locations called inducing points are selected to approximate the posterior distribution of the model. This is done to reduce computational complexity and improve model efficiency. However, inferring the posterior distribution of inducing points is not straightforward. Traditional variational inference approaches to posterior approximation often lead to significant bias. To address this issue, we propose an alternative method called Denoising Diffusion Variational Inference (DDVI) that uses a denoising diffusion stochastic differential equation (SDE) to generate  posterior samples of inducing variables. We rely on score matching methods for denoising diffusion model to approximate score functions with a neural network. Furthermore, by combining classical mathematical theory of SDEs with the minimization of KL divergence between the approximate and true processes, we propose a novel explicit variational lower bound for the marginal likelihood function of DGP. Through experiments on various datasets and comparisons with baseline methods, we empirically demonstrate the effectiveness of DDVI for posterior inference of inducing points for DGP models.
	\end{abstract}

	\section{Introduction}
	Deep Gaussian Processes (DGPs) \cite{damianou2013deep} have emerged as a robust framework for Bayesian deep learning \cite{fortuin2022priors} that allows for flexible modeling of complex functions. DGPs extend the idea of Gaussian Processes (GPs) \cite{rasmussen2003gaussian} to multiple layers, enabling the modeling of hierarchical structures and capturing intricate dependencies within the data. A crucial aspect of DGPs is the selection of inducing variables \cite{Titsias09,Snelson06,candela05}, which are sparse integration locations used to approximate the posterior distribution of the model. By leveraging these inducing points, DGPs can efficiently handle large datasets and reduce the computational burden.
	
	Variational inference  methods \cite{blei2017variational, zhang2018advances} aim to approximate the true posterior distribution with a parameterized variational distribution by minimizing their KL divergence. In the context of DGPs, traditional variational methods include mean-field Gaussian variational inference (DSVI) \cite{salimbeni2017doubly} and Implicit Posterior Variational Inference (IPVI) \cite{yu2019implicit}. However, both of these methods have their limitations and can introduce significant bias when learning the posterior distribution of inducing points.
	
	DSVI approximates the posterior distribution of inducing points with a simple Gaussian distribution. Although this approximation is analytically tractable, it often leads to substantial bias when dealing with nonlinear likelihood functions. The simplifying assumptions made in the mean-field approximation can fail to capture the complex dependencies and correlations between the inducing points, resulting in suboptimal results. On the other hand, IPVI uses a neural network to parameterize the posterior distribution of inducing points. Posterior inference is formulated as a  Nash equilibrium  \cite{awerbuch2008fast} similar to that of generative adversarial networks (GANs) \cite{goodfellow2014generative}, requiring adversarial learning for the max-max problem. However, optimizing this objective function can be challenging, especially when dealing with non-convex neural networks, and lead to instability during training and contribute to significant bias in the posterior inference of inducing points \cite{jenni2019stabilizing}.
	
	These limitations of traditional variational methods for inference of DGPs inspires the exploration of alternative approaches. Motivated by the success of denoising diffusion models \cite{rombach2022high} in deep learning, we propose a Denoising Diffusion Variational Inference (DDVI) method that utilizes the denoising diffusion SDE and incorporates principles similar to the score matching method \cite{song2020score} in order to construct the objective function.
	
	By employing the denoising diffusion SDE, we can accurately capture the complex dependencies and correlations among the inducing points. Additionally, similar to the score matching method, we can approximate the intricate score functions required for accurate posterior inference using a neural network. This combination finally allows us to explicitly derive a variational lower bound for the marginal likelihood function by KL divergence minimization, thereby addressing the bias introduced by traditional variational methods. Furthermore, DDVI incorporates numerous unique insights, including the well-developed mathematical theory of SDEs \cite{anderson1982reverse,haussmann1986time},  the bridge process trick,  stochastic optimization techniques, reparameterization techniques, and gradient backpropagation. These collectively enable us to efficiently obtain posterior samples from the denoising diffusion network. As a result, our approach improves not only the computational efficiency but also ensures stable and reliable training in DGPs.
	
	In summary, our contributions can be outlined as follows: 
	\begin{itemize}
		\item  We propose a novel parameterization approach for the posterior distribution of inducing points in DGPs, utilizing a denoising diffusion process. This method not only guarantees model efficiency by accurately capturing the complex dependencies and correlations among the inducing points, but also facilitates optimization and training.
		\item  We exploit the minimization of KL divergence between the approximate and true processes to derive an explicit variational lower bound. To efficiently obtain posterior samples, we employ stochastic optimization and reparameterization techniques for gradient backpropagation within the denoising diffusion network.
		\item Through extensive experiments on various datasets and comparisons
		with baseline methods, we empirically demonstrate the effectiveness of the DDVI method in
		posterior inference of inducing points for DGP
		models.
	\end{itemize}
	
	\section{Method}
	
	\subsection{Model Review}
	\subsubsection{Gaussian Process}
	Consider a random function \( f: \mathbb{R}^D \rightarrow \mathbb{R} \) that maps \( N \) training inputs \( \mathbf{X} \triangleq \{\mathbf{x}_n\}_{n=1}^{N} \) to a set of noisy observed outputs \( \mathbf{y} \triangleq \{ y_n \}_{n=1}^{N} \). Often, a zero mean Gaussian Process (GP) prior is assumed for the function, \( f \sim \mathcal{GP}(0, k) \), where \( k \) denotes the covariance kernel function \( k: \mathbb{R}^D \times \mathbb{R}^D \rightarrow \mathbb{R} \). Let \( \mathbf{f} \triangleq (f(\mathbf{x}_1), \ldots, f(\mathbf{x}_N))^\top \) represent the latent function values at the inputs \( \mathbf{X} \). The GP prior assumption then induces a multivariate Gaussian prior over the function values, expressed as \( p(\mathbf{f}) = \mathcal{N}(\mathbf{f} | \mathbf{0}, \mathbf{K}_{\mathbf{X}\mathbf{X}}) \), where the covariance matrix \( \mathbf{K}_{\mathbf{XX}} \) is defined by \( \left[ \mathbf{K}_{\mathbf{XX}} \right] _{ij} = k\left( \mathbf{x}_i, \mathbf{x}_j \right) \). The observed outputs \( \mathbf{y} \) are then assumed to be contaminated by i.i.d. noise, modeled as \( p(\mathbf{y} | \mathbf{f}) = \mathcal{N}(\mathbf{y} | \mathbf{f}, \sigma^2 \mathbf{I}) \), where \( \sigma^2 \) is the noise variance. The GP posterior distribution of the latent output \( p\left( \mathbf{f} | \mathbf{y} \right) \) has a closed-form solution. However, the computational cost is \( \mathcal{O}(N^3) \) and the storage requirement is \( \mathcal{O}(N^2) \), making it challenging to scale to large datasets without introducing additional techniques.
	
	Sparse methods have been developed that introduce \emph{inducing points} \(\mathbf{Z} = \{\mathbf{z}_m\}_{m=1}^{M}\) from the input space, along with corresponding \emph{inducing variables}: \(\mathbf{u} = \{f(\mathbf{z}_m)\}_{m=1}^{M}\). These methods reduce the computational complexity to \(\mathcal{O}(NM^2)\). In the \emph{Sparse Gaussian Processes} (SGPs) framework, the inducing variables \(\mathbf{u}\) and the function values \(\mathbf{f}\) share a joint multivariate Gaussian distribution, expressed as \( p(\mathbf{f}, \mathbf{u}) = p(\mathbf{f}|\mathbf{u})p(\mathbf{u}) \), with the conditional distribution given by
	\begin{equation} 
		\label{condi}
		p(\mathbf{f}|\mathbf{u}) = \mathcal{N}(\mathbf{f}|\mathbf{K}_{\mathbf{X}\mathbf{Z}}\mathbf{K}_{\mathbf{Z}\mathbf{Z}}^{-1}\mathbf{u}, \mathbf{K}_{\mathbf{X}\mathbf{X}} - \mathbf{K}_{\mathbf{X}\mathbf{Z}}\mathbf{K}_{\mathbf{Z}\mathbf{Z}}^{-1}\mathbf{K}_{\mathbf{Z}\mathbf{X}})
	\end{equation}
	and $p\left( \mathbf{u} \right) =\mathcal{N} \left(\mathbf{u}| \mathbf{0},\mathbf{K}_{\mathbf{ZZ}} \right) $ is the prior over the outputs of the inducing points. 
	
	\subsubsection{Deep Gaussian Processes}
	
	A multi-layer Deep Gaussian Process (DGP) model is a hierarchical composition of Gaussian Process (GP) models constructed by stacking multiple-output Sparse GPs (SGPs) together, as described in \cite{damianou2013deep}. Consider a DGP model with \( L \) layers, where each layer \( \ell = 1, \dots, L \) consists of \( D_\ell \) independent random functions. The output of the \((\ell-1)^\mathrm{th}\) layer, denoted as \(\mathbf{F}_{\ell-1}\), serves as the input to the \(\ell^\mathrm{th}\) layer. 
	Formally, the outputs of the \(\ell^\mathrm{th}\) layer are defined as
	\(\mathbf{F}_{\ell} \triangleq \{f_{\ell,1}(\mathbf{F}_{\ell-1}), \cdots, f_{\ell, D_{\ell}}(\mathbf{F}_{\ell-1}) \},
	\) where \( f_{\ell,d} \sim \mathcal{GP}(0, k_{\ell}) \) for \( d = 1, \dots, D_\ell \), and \(\mathbf{F}_0 \triangleq \mathbf{X} \). The inducing points and their corresponding inducing variables for each layer are denoted by \(\boldsymbol{\mathcal{Z}} \triangleq \{\mathbf{Z}_{\ell}\}_{\ell=1}^L\) and \(\mathcal{U} \triangleq \{\mathbf{U}_{\ell}\}_{\ell=1}^L\), respectively. Here, \(\mathbf{U}_{\ell} \triangleq \{f_{\ell,1}(\mathbf{Z}_{\ell}), \cdots, f_{\ell,D_{\ell}}(\mathbf{Z}_{\ell}) \}\). Let \(\mathcal{F} \triangleq \{\mathbf{F}_{\ell}\}_{\ell=1}^L\). The design of the DGP model leads to the following joint model density,
	\begin{equation}
		\label{dgplikelihood}
		p(\mathbf{y},\mathbf{F} ,\mathbf{U})=p\left( \mathbf{y}|\mathbf{F}_L \right) \prod_{\ell =1}^L{p}(\mathbf{F}_{\ell}|\mathbf{F}_{\ell -1},\mathbf{U}_{\ell})p\left( \mathbf{U} \right). 
	\end{equation}
	Here we place independent GP priors within and across layers on $\mathbf{U}$, 
	\begin{equation}
		p( \mathbf{U} ) =\prod_{l=1}^L{p( \mathbf{U}_l )}=\prod_{l=1}^L{\prod_{d=1}^{D}{\mathcal{N} \left(\mathbf{U}_{\ell,d}| 0,\mathbf{K}_{\mathbf{Z}_{\ell}\mathbf{Z}_{\ell}} \right)}}
	\end{equation}
	and the condition similar to Eq. (\ref{condi}) is defined as follows,
	\begin{equation}
		\label{condid}
		p\left(\mathbf{F}_{\ell} \mid \mathbf{F}_{\ell-1}, \mathbf{U}_{\ell}\right) =\prod_{d=1}^{D_{\ell}} \mathcal{N}\left(\mathbf{F}_{\ell, d} \mid \mathbf{\mu}_{\ell,d} , \mathbf{\Sigma}_{\ell,d}\right),
	\end{equation}
	where we define
	\begin{eqnarray}
		\mathbf{\mu}_{\ell,d} & =& \mathbf{K}_{\mathbf{F}_{\ell-1}\mathbf{Z}_{\ell}} \mathbf{K}_{\mathbf{Z}_{\ell} \mathbf{Z}_{\ell}}^{-1} \mathbf{U}_{\ell, d}\nonumber \\
		\mathbf{\Sigma}_{\ell,d} & =& \mathbf{K}_{\mathbf{F}_{\ell-1}\mathbf{F}_{\ell-1}} -\mathbf{K}_{\mathbf{F}_{\ell-1}\mathbf{z}_{\ell} } \mathbf{K}_{\mathbf{Z}_{\ell} \mathbf{Z}_{\ell}}^{-1} \mathbf{K}_{\mathbf{Z}_{\ell} \mathbf{F}_{\ell-1}}. \nonumber
	\end{eqnarray}
	
	In the context of DGPs, traditional variational methods primarily include mean-field Gaussian variational inference (DSVI) \cite{salimbeni2017doubly} and Implicit Posterior Variational Inference (IPVI) \cite{yu2019implicit}. However, both of these methods have their limitations and can introduce significant bias in inferring the posterior distribution of inducing points.
	
	DSVI approximates the posterior by a mean-field Gaussian, $q\left( \mathbf{U}_{\ell ,1:D_{\ell}} \right) =\mathcal{N} \left(\mathbf{U}_{\ell ,1:D_{\ell}}\mid 
	\mathbf{m}_{\ell, 1:D_{\ell}},\mathbf{S}_{\ell ,1:D_{\ell}} \right) $, where $\mathbf{m}_{\ell ,1:D_{\ell}}$ and $\mathbf{S}_{\ell ,1:D_{\ell}}$ are variational parameters. However, this assumption is overly strict and may limit the effectiveness and expressiveness of the model. The likelihood $p(\mathbf{y}|\mathbf{U})$  is difficult to compute because the latent functions $\mathbf{F}_1,\cdots, \mathbf{F}_{L-1}$ are  all non-linear kernel functions. 
	
	On the other hand, IPVI utilizes a neural network $\xi$ to parameterize the posterior distribution of inducing points. Posterior inference is formulated as a Nash equilibrium  \cite{awerbuch2008fast} similar to that of generative adversarial networks (GANs) \cite{goodfellow2014generative}, requiring adversarial learning for the max-max problem, 
	\begin{equation}
		\begin{aligned}
			l_\mathrm{IPVI}(\xi)&= \mathbb{E}_{q_{\xi}(\mathbf{U})}\left[\log p(\mathbf{y}|  \mathbf{U})-D_{\psi^\star}(\mathbf{U})\right] \\
			s.t. \quad \psi^\star&=\mathop {\mathrm{arg}\max} \limits_{\psi}\, \mathbb{E}_{p(\mathbf{U})}[\log \left(1-\sigma\left(D_{\psi}(\mathbf{U})\right)\right]\\&+\mathbb{E}_{q_{\xi}(\mathbf{U})}[\log \sigma\left(D_{\psi}(\mathbf{U})\right)],
		\end{aligned}
	\end{equation}
	where $D_{\psi}$ is another discriminator network. However, optimizing such an implict objective $l(\xi)$ can be challenging, especially when dealing with the non-convex neural network $D_{\psi}$. This can lead to instability during training and contribute to significant bias in the posterior inference of inducing points \cite{jenni2019stabilizing}.
	
	To address this issue, we propose a novel parameterization approach for the
	posterior distribution of inducing variables $\mathbf{U}$ that uses a denoising diffusion process. This method not only ensures model efficiency by accurately capturing the complex dependencies and correlations among the inducing points, but also facilitates optimization and training.

	\subsection{Denoising Diffusion Variational Inference}
	\subsubsection{Parameterizing Inducing Point Posteriors}
	
	Let $H=D\times M\times L$ denote the dimension of the inducing points. We aim to sample from the true posterior distribution $q(\mathbf{U})$ in $\mathbb{R}^{H}$, $q(\mathbf{U})=p(\mathbf{U}|\mathbf{y})$. Following a similar setup to prior works \cite{tzen2019theoretical,zhang2021path,vargas2023denoising}, we start by sampling from a fixed distribution $p_{\mathrm{fix}}$ and then follow a Markov process in which we consider a sequential latent variable model with a joint distribution denoted as  $\mathcal{Q}\left(\mathbf{U}_0, \ldots, \mathbf{U}_T\right)$ , for step $t_s \in \{0,..,T-1\}$,
	\begin{equation}
		\mathbf{U}_{t_s+1} \sim \mathcal{T}\left(\mathbf{U}_{t_s+1} \mid \mathbf{U}_{t_s}\right), \qquad \mathbf{U}_0 \sim p_ \mathrm{fix}   
	\end{equation}
	Here $\mathcal{T}\left(\mathbf{U}_{t_s+1} \mid \mathbf{U}_{t_s}\right)$ denotes a transition probability distribution. Through this sequence model, we use the marginal distribution $\mathcal{Q}\left(\mathbf{U}_T\right)$ at the terminal step $T$ to approximate the true posterior distribution $q\left(\mathbf{U}_T\right)$.

	\subsubsection{Time-reversal Representation of Diffusion SDE}
	In this paper, we constrain the Markov process $\mathcal{Q}\left(\mathbf{U}_0, \ldots, \mathbf{U}_T\right)$ to be a  time-reversal process of the following  forward noising diffusion stochastic differential equation (SDE),
	\begin{equation}
		\label{eq2}
		\mathrm{d}\overrightarrow{\mathbf{U}}_t=\mathbf{h}(t, \overrightarrow{\mathbf{U}}_t)\mathrm{d}t+g(t)\mathrm{d}B_t,\qquad \overrightarrow{\mathbf{U}}_0\sim q,
	\end{equation}
	
	where $\mathbf{h} (t, \cdot): \mathbb{R}^H \rightarrow \mathbb{R}^H$ is the drift coefficient, $g(t) \in \mathbb{R}$ is the diffusion coefficient, and $\left(B_t\right)_{t \in[0, T]}$ is an $H$-dimensional Brownian motion. This diffusion induces the path measure $\mathcal{P}$ on the time interval $[0, T]$ and the marginal density of $\overrightarrow{\mathbf{U}}_t$ is denoted $p_t$. Note that by definition we always have $p_0=q$ when using an SDE to perturb this distribution. In DDPM \cite{ho2020denoising,song2020score}, $p_T$ is an unstructured prior distribution that contains no information of $p_0$, such as a
	Gaussian distribution with fixed mean and variance.
	
	From \cite{anderson1982reverse,haussmann1986time}, the time-reversal representation of Eq. (\ref{eq2}),   $\overleftarrow{\mathbf{U}}_t=\overrightarrow{\mathbf{U}}_{T-t}$, where equality is here in distribution, satisfies
	\begin{eqnarray}
		\label{eq3}
		\mathrm{d} \overleftarrow{\mathbf{U}}_t & \hspace{-10pt}=\hspace{-10pt} & g(T-t)^2 \nabla  \ln \big(p_{T-t}(\overleftarrow{\mathbf{U}}_t)\big)\mathrm{~d}t -\mathbf{h}(T-t, \overleftarrow{ \mathbf{U}}_t) \mathrm{~d}t \nonumber \\ 
		&& +g(T-t) \mathrm{~d} W_t, \nonumber\\ 
		\overleftarrow{\mathbf{U}}_0 & \hspace{-10pt}\sim \hspace{-10pt} &  p_T,
	\end{eqnarray}
	where $\left(W_t\right)_{t \in[0, T]}$ is another $H$-dimensional Brownian motion. By definition, this time-reversal starts from $$\overleftarrow{\mathbf{U}}_0 \sim p_T \approx p_\mathrm{fix}$$ and is such that $\overleftarrow{\mathbf{U}}_T \sim q$. Since the distribution of $\overleftarrow{\mathbf{U}}_T$ is consistent with the true posterior $q$ , we can parameterize the transition probability $\mathcal{T}\left(\mathbf{U}_{t_s+1} \mid \mathbf{U}_{t_s}\right)$ in the Euler discretized form of Eq. (\ref{eq3}).
	
	\subsubsection{Score Matching Technique}
	This suggests that if we could approximately simulate the diffusion of Eq. (\ref{eq3}), then we could obtain approximate samples from the target $q$. However, putting this idea in practice requires being able to approximate the intractable scores $\nabla\ln \big(p_t(\cdot)\big)$ for $t \in[0, T]$. To achieve this, DDPM \cite{ho2020denoising,song2020score} rely on score matching techniques. Specially, to approximate $\mathcal{P}$ consider a path measure $\mathcal{P}^\phi$  whose time-reversal is induced by
	\begin{eqnarray}
		\label{eq5}
		\mathrm{d} \overleftarrow{\mathbf{U}}_t^\phi & \hspace{-12pt}=\hspace{-12pt} & g(T-t)^2 s_\phi(T-t, \overleftarrow{\mathbf{U}}_t^\phi) \mathrm{d}t-\mathbf{h}(T-t, \overleftarrow{\mathbf{U}}_t^\phi)\mathrm{d}t\nonumber\\
		& & + ~ g(T-t) \mathrm{~d} W_t,\nonumber\\
		\overleftarrow{\mathbf{U}}_0^\phi &\hspace{-12pt}\sim\hspace{-12pt}& p_\mathrm{fix},     
	\end{eqnarray}
	so that the backward process $\overleftarrow{\mathbf{U}}_t^\phi \sim \mathcal{Q}_{t}^\phi$, where $p_\mathrm{fix}$ represents a fixed distribution. To obtain $s_\phi(t, \cdot) \approx \nabla \ln \big(p_t(\cdot)\big)$, we parameterize $s_\phi(t, \cdot)$ by a neural network whose parameters are obtained by minimizing $\mathrm{KL}(\mathcal{P}||\mathcal{P}^\phi)$. From the chain rule for the KL divergence \cite{leonard2013survey} we  have,
	\begin{equation}
		\label{eq6}
		\mathrm{KL}(\mathcal{P}||\mathcal{P}^\phi) = \mathrm{KL}(p_T||p_\mathrm{fix})+\mathrm{KL}(\mathcal{P}\cdot|\overrightarrow{\mathbf{U}}_T)||\mathcal{P}^\phi(\cdot|\overrightarrow{\mathbf{U}}_T^\phi))
	\end{equation}
	where by the well-known Girsanov Theorem \cite{oksendal2013stochastic} and the martingale property of Itô integrals the second term on the RHS is
	\begin{eqnarray}
		\mathrm{KL}(\mathcal{P}(\cdot|\overrightarrow{\mathbf{U}}_T)||\mathcal{P}^\phi(\cdot|\overrightarrow{\mathbf{U}}_T^\phi)) = \hspace{1.2in}&&\\
		&&\nonumber\\
		\frac{1}{2} \int_0^T \mathbb{E}_{\mathcal{P}_t}\left[g(t)^2\left\|\nabla \ln\big( p_t(\overrightarrow{\mathbf{U}}_t)\big)-\boldsymbol{s}_{\boldsymbol{\phi}}(\overrightarrow{\mathbf{U}}_t, t)\right\|_2^2\right] \mathrm{d}t && \nonumber 
	\end{eqnarray}
	From the denoising score matching derivation \cite{vincent2011connection}, this can also be written as
	$$\frac{1}{2} \int_0^T \mathbb{E}\left[g(t)^2\|\nabla \ln\big( p_t(\overrightarrow{\mathbf{U}}_t|\overrightarrow{\mathbf{U}}_0)\big) -\boldsymbol{s}_{\boldsymbol{\phi}}(\overrightarrow{\mathbf{U}}_t, t)\|_2^2\right] \mathrm{d} t$$
	plus a constant term, where the expectation is over the joint distribution $p_0(\overrightarrow{\mathbf{U}}_0)p_t(\overrightarrow{\mathbf{U}}_t|\overrightarrow{\mathbf{U}}_0)$.
	
	As the main loss function in DDPM, diffusion-based generative modeling approaches typically rely on Eq. (\ref{eq6}), which involves sampling from $p_0$, the original data such as images, and then backpropagating to estimate the parameters of the neural network $s_\phi$. However, unlike traditional score matching techniques, this loss function is not applicable to our model since our $p_0$ is the posterior probability $q$ and we only have access to the joint likelihood, and cannot sample from it. To address this issue, we propose an alternative approach by minimizing  $\mathrm{KL}(\mathcal{P}^\phi||\mathcal{P})$. Analogous to Eq. (\ref{eq6}), considering that we can only obtain samples from $\mathcal{Q}_{t}^\phi$, we have,
	\begin{eqnarray}
		\mathrm{KL}(\mathcal{P}^\phi||\mathcal{P}) &\hspace{-5pt}=\hspace{-5pt}& \mathrm{KL}(\mathcal{Q}^\phi||\mathcal{Q})\\
		&\hspace{-5pt}=\hspace{-5pt}&\mathrm{KL}(p_\mathrm{fix}||p_T)+\mathrm{KL}(\mathcal{Q}^\phi(\cdot|\overleftarrow{\mathbf{U}}_0^\phi)||\mathcal{Q}(\cdot|\overleftarrow{\mathbf{U}}_0))\nonumber
	\end{eqnarray}
	where
	\begin{equation}
		\label{eq7}
		\mathrm{KL}(\mathcal{Q}^\phi(\cdot|\overleftarrow{\mathbf{U}}_0^\phi)||\mathcal{Q}(\cdot|\overleftarrow{\mathbf{U}}_0)) =
		\frac{1}{2} \int_0^T \mathbb{E}\,g(T-t)^2\|\varphi(t)\|_2^2 \mathrm{d}t
	\end{equation}
	where the expectation is over $\mathcal{Q}_t^\phi$ and we have defined
	$$\varphi(t) \triangleq \nabla \ln \big(p_{T-t}(\overleftarrow{\mathbf{U}}_t^\phi)\big) -\boldsymbol{s}_{\boldsymbol{\phi}}(T-t, \overleftarrow{\mathbf{U}}_t^\phi) $$
	
	However, the current challenge we face is that, although we can obtain samples from $\mathcal{Q}_{t}^\phi$ by simulating the SDE (\ref{eq5}), dealing with the nonlinear drift function of SDE (\ref{eq5}) makes it difficult to obtain $\nabla \ln \big( p_{T-t}(\overleftarrow{\mathbf{U}}_t^\phi)\big)$ in Eq. (\ref{eq7}).
	
	\subsubsection{Bridge Process Trick}
	Therefore, we propose an alternative approach by constructing a bridge process $\mathcal{P}^\mathrm{Bri}$ to  assist in measuring   $\mathrm{KL}(\mathcal{P}^\phi||\mathcal{P})$. First, we observe that 
	\begin{eqnarray}
		\label{eq8}
		\mathrm{KL(}\mathcal{P} ^{\phi}||\mathcal{P} ) &=& \mathbb{E} _{\mathcal{P} ^{\phi}}\log \frac{\mathrm{d}\mathcal{P} ^{\phi}}{\mathrm{d}\mathcal{P}}\\
		&=&\mathbb{E} _{\mathcal{P} ^{\phi}}\log \frac{\mathrm{d}\mathcal{P} ^{\phi}}{\mathrm{d}\mathcal{P} ^{\mathrm{Bri}}}+\mathbb{E} _{\mathcal{P} ^{\phi}}\log \frac{\mathrm{d}\mathcal{P} ^{\mathrm{Bri}}}{\mathrm{d}\mathcal{P}}\nonumber
	\end{eqnarray}
	where we represent the stochastic process $\mathrm{KL}$ with the Radon-Nikodym derivative. Given the specific form in Eq. (\ref{eq8}), we define the bridge process $\mathcal{P}^\mathrm{Bri}$ to follow the diffusion formula as in Eq. (\ref{eq2}), but initialized at $p_0^{\mathrm{Bri}}(\overrightarrow{\mathbf{U}}_0^\mathrm{Bri}) = p_\mathrm{fix}$ instead of $q$, which aligns with the distribution of $\overleftarrow{\mathbf{U}}_0$ in Eq. (\ref{eq5}),
	\begin{equation}
		\label{eq9}
		\mathrm{d}\overrightarrow{\mathbf{U}}_t^\mathrm{Bri}=\mathbf{h}(t, \overrightarrow{ \mathbf{U}}_t^\mathrm{Bri})\mathrm{d}t+g(t)\mathrm{d}B_t,\quad \overrightarrow{\mathbf{U}}_0^\mathrm{Bri}\sim p_\mathrm{fix}.
	\end{equation}
	We typically assume $\mathbf{h}(\cdot, t)$ is affine, $
	\mathbf{h}(x,t)=-\mathbf{\lambda }\left( t \right) x
	$ and $p_\mathrm{fix}=\mathcal{N}(0,\sigma^2 I)$. Then the transition kernel $p_t(\overrightarrow{\mathbf{U}}_t^\mathrm{Bri}|\overrightarrow{\mathbf{U}}_0^\mathrm{Bri})$ is always a Gaussian distribution $\mathcal{N}(l_t,\Sigma_t)$, where the mean $l_t$ and variance $\Sigma_t$  are often known in closed-forms \cite{sarkka2019applied} by 
	\begin{equation}
		\label{eq10}
		\begin{aligned}
			\frac{\mathrm{d} l_t}{\mathrm{d}t} & =-\lambda (t) l_t, \qquad\qquad\qquad\quad\, l_0=0 \\
			\frac{\mathrm{d} \Sigma_t}{\mathrm{d}t} & =-2 \lambda (t) \Sigma_t+g(t)^2I, \qquad \Sigma_0=\sigma^2 I
		\end{aligned}   
	\end{equation}
	By the calculations of ordinary differential equations \cite{hale2013introduction}, we obtain the following general solution to Eq. (\ref{eq10}),
	\begin{equation}
		\label{eq11}
		\begin{aligned}
			l_t&=l_0 e^{-\int_0^t \lambda(s) \mathrm{d} s},\\
			\Sigma_t&=\left(\int_0^t g(r)^2e^{\int_0^r  \lambda(s) \mathrm{d} s}  \mathrm{d} rI+\Sigma_0\right)e^{-\int_0^t  \lambda(s) \mathrm{d} s}
		\end{aligned}
	\end{equation}
	According to Eq. (\ref{eq11}) we can derive from the Gaussian linear transformation principle that for any $t$, the distribution $p_t^{\mathrm{Bri}}$ of $\overrightarrow{\mathbf{U}}_t^\mathrm{Bri}$ is a zero-mean Gaussian distribution,
	\begin{equation}
		\label{eq12}
		\begin{aligned}
			p_t^{\mathrm{Bri}}(\overrightarrow{\mathbf{U}}_t^\mathrm{Bri})&=\int p_t(\overrightarrow{\mathbf{U}}_t^\mathrm{Bri}|\overrightarrow{\mathbf{U}}_0^\mathrm{Bri}) p_t(\overrightarrow{\mathbf{U}}_0^\mathrm{Bri})d\overrightarrow{\mathbf{U}}_0^\mathrm{Bri}  \\
			&=\mathcal{N}(0,\kappa _tI)
		\end{aligned}
	\end{equation}
	where we have defined the variance 
	$$\kappa _t \triangleq \left(\int_0^t g(r)^2e^{\int_0^r  \lambda(s) \mathrm{d} s}  \mathrm{d} r+\sigma^2\right)e^{-\int_0^t  \lambda(s) \mathrm{d} s}.$$ 
	
	We can write the SDE equation for the reverse process $\mathcal{Q}^\mathrm{Bri}$ of $\mathcal{P}^\mathrm{Bri}$ as
	\begin{eqnarray}
		\label{eq13}
		\mathrm{d} \overleftarrow{\mathbf{U}}_t^\mathrm{Bri} &=& g(T-t)^2 \nabla \ln \big( p_{T-t}^\mathrm{Bri}\big(\overleftarrow{\mathbf{U}}_t^\mathrm{Bri}\big)\big) \mathrm{d} t\nonumber\\
		&&-~\mathbf{h}(T-t,\overleftarrow{\mathbf{U}}_t^\mathrm{Bri})\mathrm{d} t\nonumber \\
		&&+~g(T-t) \mathrm{~d} W_t,\\
		\overleftarrow{\mathbf{U}}_0^\mathrm{Bri} &\sim& p_T^\mathrm{Bri}.\nonumber     
	\end{eqnarray}
	According to Eq. (\ref{eq12}), we can obtain an analytical expression for the derivative of the log-likelihood function with respect to $\overleftarrow{\mathbf{U}}_t^\mathrm{Bri}$,
	\begin{equation}
		\label{eq14}
		\nabla \ln \big(p_{T-t}^\mathrm{Bri}\big(\overleftarrow{\mathbf{U}}_t^\mathrm{Bri}\big)\big)=-\frac{\overleftarrow{\mathbf{U}}_t^\mathrm{Bri}}{\kappa _{T-t}}.
	\end{equation}
	
	Next we calculate the value of Eq. (\ref{eq8}). For the first term, according to the chain rule for KL and Girsanov Theorem \cite{oksendal2013stochastic}, incorporating Eqs. (\ref{eq5}, \ref{eq13}, \ref{eq14}), we have
	$$\mathbb{E} _{\mathcal{P} ^{\phi}}\log \frac{\mathrm{d}\mathcal{P} ^{\phi}}{\mathrm{d}\mathcal{P}^{\mathrm{Bri }}}=\mathrm{KL}\left(\mathcal{P}^\phi \| \mathcal{P}^{\mathrm{Bri }}\right)=\mathrm{KL}\left(\mathcal{Q}^\phi \| \mathcal{Q}^{\mathrm{Bri }}\right)$$
	which can be broken down into the sum of two terms,
	\begin{equation}\label{eq15}
		\operatorname{KL}\left(p_\mathrm{fix} \| p_T^{\mathrm {Bri }}\right)+\mathrm{KL}(\mathcal{Q}^\phi(\cdot|\overleftarrow{\mathbf{U}}_0^\phi)||\mathcal{Q}(\cdot|\overleftarrow{\mathbf{U}}_0^\mathrm{Bri}))
	\end{equation}
	where
	$$\mathrm{KL}(\mathcal{Q}^\phi(\cdot|\overleftarrow{\mathbf{U}}_0^\phi)||\mathcal{Q}(\cdot|\overleftarrow{\mathbf{U}}_0^\mathrm{Bri})) =\hspace{1.5in}$$
	$$\frac{1}{2} \int_0^T \mathbb{E}_{\mathcal{Q}_t^\phi}~ g(T-t)^2\Big\|\frac{\overleftarrow{\mathbf{U}}_t^\mathrm{\phi}}{\kappa _{T-t}}+\boldsymbol{s}_{\boldsymbol{\phi}}(T-t, \overleftarrow{\mathbf{U}}_t^\phi)\Big\|_2^2~ \mathrm{d}t$$
	
	At this point, we can simulate the SDE (\ref{eq5}) to compute the first term in Eq. (\ref{eq8}).  The integral term can be computed using either ODE solvers \cite{chen2018neural} or by employing Riemann summation methods. For the second term, $\mathbb{E} _{\mathcal{P} ^{\phi}}\log \frac{\mathrm{d}\mathcal{P} ^{\mathrm{Bri}}}{\mathrm{d}\mathcal{P}}$, we can see from Eq. (\ref{eq2}) and Eq. (\ref{eq9}) that $\mathcal{P}$ and $\mathcal{P}^{\mathrm{Bri}}$ have the same dynamic system $\tau$, except for different initial values. Therefore, we have
	\begin{eqnarray}
		\label{eq16}
		\mathbb{E} _{\mathcal{P} ^{\phi}}\log \frac{\mathrm{d}\mathcal{P} ^{\mathrm{Bri}}}{\mathrm{d}\mathcal{P}} & = & \mathbb{E} _{\mathcal{P} ^{\phi}}\log \frac{\mathcal{P} ^{\mathrm{Bri}}\left( \tau |\cdot \right) p_{0}^{\mathrm{Bri}}\left( \cdot \right)}{\mathcal{P} \left( \tau |\cdot \right) p_0\left( \cdot \right)}\nonumber\\
		&=&\mathbb{E} _{\mathcal{Q}_T ^{\phi}}\log \frac{p_{0}^{\mathrm{Bri}}\left( \cdot \right)}{p_0\left( \cdot \right)}\\
		&=&\mathbb{E} _{\mathcal{Q}_T ^{\phi}}\log \frac{p_{\mathrm{fix}}}{q}\nonumber
		\\
		&=&\mathbb{E} _{\mathcal{Q}_T ^{\phi}}\log \frac{p_{\mathrm{fix}}}{p(\mathbf{y}|\cdot)p(\cdot)}+\log p(\mathbf{y})\nonumber
	\end{eqnarray}
	\subsubsection{A New Evidence Lower Bound}
	Let $l_1(\phi)=\mathbb{E} _{\mathcal{P} ^{\phi}}\log \frac{\mathrm{d}\mathcal{P} ^{\phi}}{\mathrm{d}\mathcal{P}^{\mathrm{Bri }}}$. Combining Eqs. (\ref{dgplikelihood}, \ref{eq8}, \ref{eq15}, \ref{eq16}), we obtain a new variational lower bound $l(\phi)$ for the marginal likelihood $\log p(\mathbf{y})$ of our method,
	\begin{equation}
		\label{eq17}
		\begin{aligned}
			\log p(\mathbf{y})= ~& \mathrm{KL(}\mathcal{P} ^{\phi}||\mathcal{P} )-l_1(\phi)-\mathbb{E} _{\mathcal{Q}_T ^{\phi}}\log \frac{p_{\mathrm{fix}}}{p(\mathbf{y}|\mathbf{\cdot})p(\mathbf{\cdot})}\\
			=~& \mathrm{KL(}\mathcal{P} ^{\phi}||\mathcal{P} )-l_1(\phi)-\mathbb{E} _{\mathcal{Q}_T ^{\phi}}\log {p_{\mathrm{fix}}}\\&+\mathbb{E} _{\mathcal{Q}_T ^{\phi}}\log p(\cdot)+\mathbb{E} _{\mathcal{Q}_T^{\phi}, \mathbf{F_1},..., \mathbf{F_{L}}}\log p(\mathbf{y}| \mathbf{F_{L}})\\
			\geqslant~ & \mathbb{E} _{\mathcal{Q}_T ^{\phi}}\log p(\cdot)+\mathbb{E} _{\mathcal{Q}_T^{\phi}, \mathbf{F_1},..., \mathbf{F_{L}}}\log p(\mathbf{y}| \mathbf{F_{L}})\\&-l_1(\phi)-\mathbb{E} _{\mathcal{Q}_T ^{\phi}}\log {p_{\mathrm{fix}}}\\
			=~&l(\phi)
		\end{aligned}
	\end{equation}
	
	In our derivation, $p(\cdot)$ represents the prior function of $\mathbf{U}$. By introducing a new variational lower bound for $\log p(\mathbf{y})$, our proposed model, compared to the initial mean-field variational inference model (DSVI) where $q$ is approximated by a Gaussian distribution, approximates the posterior distribution through a  denoising diffusion process. The flexibility of the denoising neural network $\phi$ intuitively suggests that our model has an advantage in approximating the posterior distribution. On the other hand, compared to IPVI, DDVI provides an explicit evidence lower bound (ELBO), which means it is easier to train and allows for efficient backpropagation.
	
	\subsection{Reparameterization Trick and SGD}
	\input{algorithm}
	For ease of sampling, we consider a reparameterization version of Eq. (\ref{eq17}) based on the approaximate transition probability $\mathcal{T}_\phi\left(\mathbf{U}_{t_s+1} \mid \mathbf{U}_{t_s}\right)$ given by
	\begin{align}
		\begin{aligned}
			\mathcal{T}_{\phi }(\mathbf{U}_{t_s+1})=~&\mathbf{U}_{t_s}-\mathbf{h}(\mathbf{U}_{t_s}, T-t_s) ~+\\&g(T-t_s)^2 s_\phi \left(T-t_s,\mathbf{U}_{t_s}\right)+g(T-t) \boldsymbol{\epsilon}_{t_s} 
		\end{aligned}
	\end{align}
	where $\boldsymbol{\epsilon}_{t_s} \sim \mathcal{N}(0,I)$. Given that $\mathbf{U}_{t_s+1} =\mathcal{T}_{\phi}(\mathbf{U}_{t_s})$, we have a representation of $\mathbf{U}_{t_s}$ by a stochastic flow,
	\begin{align}
		\mathbf{U}_{t_s+1}=\mathcal{T}_{\phi}\left(\mathbf{U}_{t_s}\right)=\mathcal{T}_{\phi}\circ \mathcal{T}_{\phi}\circ\cdots \mathcal{T}_{\phi}(\mathbf{U}_0).     
	\end{align}
	Moreover, for DGP models, we also have a reparameterization version \cite{salimbeni2017doubly} of the conditional distribution  in Eq. (\ref{condid}) of the form 
	\begin{align}
		\begin{aligned}
			\mathbf{F}_{\ell,d}=~&\mathbf{K}_{\mathbf{F}_{\ell-1}\mathbf{Z}_{\ell}} \mathbf{K}_{\mathbf{Z}_{\ell} \mathbf{Z}_{\ell}}^{-1} \mathbf{U}_{\ell, d}\\&+\sqrt{\mathbf{K}_{\mathbf{F}_{\ell-1}\mathbf{F}_{\ell-1}} -\mathbf{K}_{\mathbf{F}_{\ell-1}\mathbf{z}_{\ell} } \mathbf{K}_{\mathbf{Z}_{\ell} \mathbf{Z}_{\ell}}^{-1} \mathbf{K}_{\mathbf{Z}_{\ell} \mathbf{F}_{\ell-1}}}\boldsymbol{\epsilon }_{\ell,d}
		\end{aligned}
	\end{align}
	where  $\boldsymbol{\epsilon}_{\ell, d} \in \mathbb{R}^{N}$ are standard Gaussian random variables. In order to accelerate training and sampling in our inference scheme, we propose a  scalable variational bound that is tractable in the large data regime based on stochastic variational inference \cite{kingma2013auto,hoffman2015structured,salimbeni2017doubly,naesseth2020markovian} and stochastic gradient descent \cite{welling2011bayesian,chen2014stochastic,zou2019stochastic, alexos2022structured}. Specially, instead of computing  the full log likelihood, we use a stochastic variant to subsample datasets into a mini-batches $\mathcal{D}_I$ with $\left | \mathbf{X}_I\ \right | =B$, where $I\subset \{1,2,..,N\}$ is the index of
	a mini-batch. We present the resulting stochastic inference for our Denoising Diffuison Variational Inference algorithm for DGP models in Algorithm \ref{algorithm}.
	
	\subsection{Predictive Distribution}
	To obtain the final layer density for making predictions, we first sample from the optimized generator and transform the input locations $\mathbf{x}$ to the test locations $\mathbf{x}^\star$ using Eq. (\ref{dgplikelihood}). We subsequently compute the function values at the test locations, which are represented as $\mathbf{F}_{\ell}^\star$. Finally, we use  the equation below to estimate the density of the final layer, which enables us to make predictions for the test data
	$$q(\mathbf{F}_{L}^{\star})\hspace{-2pt}=\hspace{-4pt}\int{{\prod_{d,\ell}{p(\mathbf{F}_{\ell ,d}^{\star}|\mathbf{F}_{\ell -1}^{\star},\mathbf{U}_{\ell ,d})\mathcal{Q}_T ^{\phi}\left( \mathbf{U}_{\ell ,d} \right) d\mathbf{F}_{\ell -1}^{\star}d\mathbf{U}_{\ell ,d}}}}$$
	where $\mathcal{Q}_T ^{\phi}$ represents the output of the denoising diffusion process at time $T$ and the first term of the integral $p( \mathbf{F}_{\ell,d}^\star|\mathbf{F}_{\ell -1}^\star,\mathbf{U}_{\ell ,d} )$ is conditional Gaussian. We leverage this to draw samples from $q\left(\mathbf{F}_{L}^\star\right)$ and further perform the sampling according to the problem considered. 

\section{Experiments}

\begin{figure}[t]
	\begin{center}
		\centerline{\includegraphics[width=9cm]{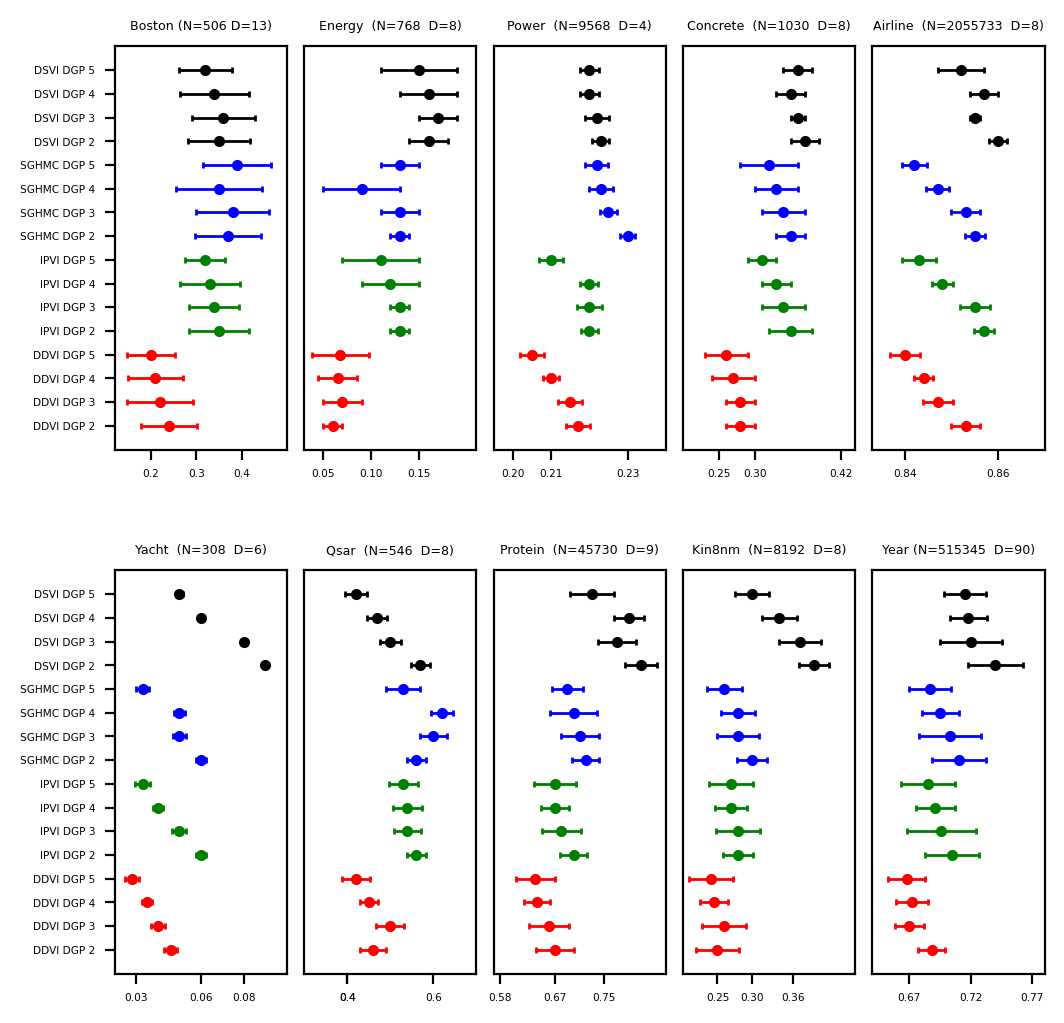}}
		\caption{Regression  test RMSE results by our DDVI method (red), SGHMC (blue), IPVI(green) and DSVI (black) for DGPs on ten UCI benchmark datasets.  The numbers 2, 3, 4, and 5 represent the layers of DGP methods. Lower is better. The mean is shown with error bars of one standard error. The dimensions of the data are displayed above each subgraph.}
		\label{fig:mse}
	\end{center}
\end{figure}
\begin{figure}[t]
	\begin{center}
		\centerline{\includegraphics[width=9cm]{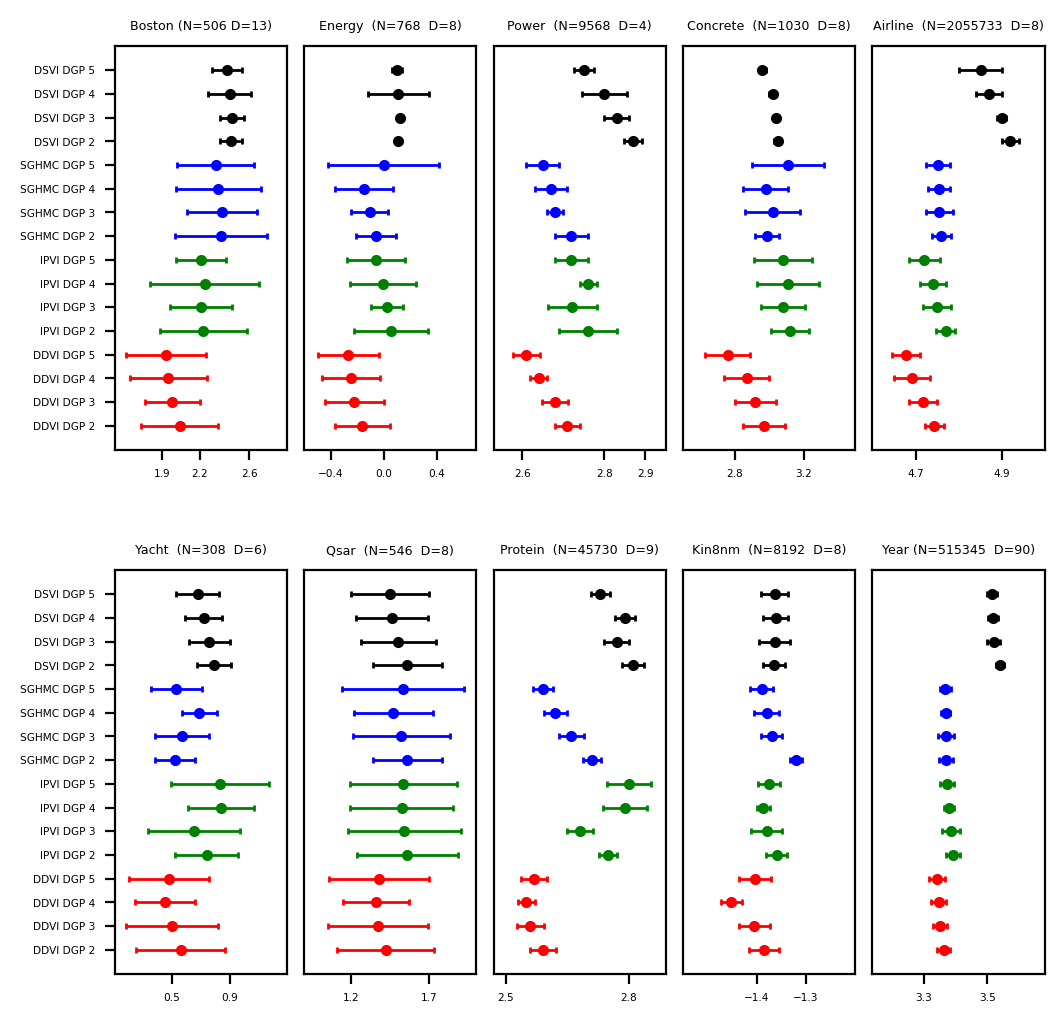}}
		\caption{Regression  test mean NLL results by our DDVI method (red), SGHMC (blue), IPVI (green) and DSVI (black) for DGPs on ten UCI benchmark datasets. The numbers 2, 3, 4, and 5 represent the layers of DGP methods. Lower is better. The mean is shown with error bars of one standard error. The dimensions of the data are displayed above each subgraph.}
		\label{fig:nll}
	\end{center}
\end{figure}

\subsection{Baseline Models and Hyperparameter Settings}
In order to evaluate the performance of our proposed method, we conducted empirical evaluations on real-world datasets for both regression and classification tasks, with both small and large datasets. We compare against several other models, including Doubly Stochastic VI (DSVI) \cite{salimbeni2017doubly}, Implicit Posterior VI (IPVI) \cite{yu2019implicit}, and the state-of-the-art SGHMC model \cite{havasi2018inference}. All experiments were conducted with the same hyper-parameters and initializations whenever possible to obtain a fair comparison.

We constructed a random $0.9 / 0.1$ train/test split and normalized the features of our datasets to the range $[-1, 1]$. The depth $L$ of DGP models varied from $2$ to $5$, with $128$ inducing points per layer, which were initialized by sampling Gaussian random variables. The output dimension for each hidden layer is set to $1$ for the final layer and the dimensionality of the data for all others. We use the RBF kernel for all tasks. For all datasets, we have optimized hyper-parameters and network parameters jointly and set learning rate to $0.01$ using Adam optimizer \cite{kingma2014adam}. We trained all models on all datasets until convergence was achieved. In each experiment, we repeated the process 10 times and reported the mean and standard deviation of the metrics. The selection of the denoising diffusion networks are done manually by  the classical grid search approach for each experimental dataset. It is worth mentioning that this work also benefits from the contributions of the PyTorch platform, GPyTorch \cite{gardner2018gpytorch}, and related work on neural SDE solvers \cite{li2020scalable,kidger2021efficient}. All our experiments were conducted on an RTX 4090 GPU.

\subsection{Regression Task}
In our experiments, we evaluated the performance of the DDVI model on ten UCI regression datasets, which varied in size from 308 to 2,055,733 data points. We used the mean RMSE and mean NLL \cite{gneiting2007strictly} of the test data as the performance metric, and the results are presented in Figure \ref{fig:mse} and Figure \ref{fig:nll}. 

As shown in these two figures, our DDVI method consistently achieves competitive results compared to three baselines on the majority of datasets. This is attributed to the key advantages of our approach, which overcomes limitations present in previous methods  as discussed in the main text. Our findings also suggest that deeper DGP models tend to perform better.  It is worth mentioning that the difference in performance may be attributed to the nature of the datasets, such as their size or the presence of outliers or singular values.

Using mini-batch algorithm and GPU acceleration, our method can also be extended to larger datasets. Our evaluation of the performance of DDVI in Figures \ref{fig:mse} and \ref{fig:nll} is also conducted on two real-world large-scale regression datasets: the YearMSD dataset and the Airline dataset. The YearMSD dataset has a large input dimension of $D = 90$ and a data size of approximately 500,000. The Airline dataset, on the other hand, has an input dimension of $D = 8$ and a large data size of approximately 2 million. For the YearMSD dataset, we split the data into training and test sets, using the first 463,810 examples as training data and the last 51,725 examples as test data. Similarly, for the Airline dataset, we take the first 700K points for training and next 100K for testing.

\subsection{Image and Large-Scale Dataset Classification}
We evaluate our method on multiclass classification tasks using the MNIST \cite{lecun1998mnist}, Fashion-MNIST \cite{xiao2017fashion}, and CIFAR-10 \cite{krizhevsky2009learning} datasets. The first two datasets consist of grayscale images of size $28\times28$ pixels, while CIFAR-10 comprises colored images of size $32\times32$ pixels. The results are presented in Table \ref{tab:image-classification}. We note that our method outperforms the other three methods on all three datasets, with significantly less training time. Specially, for CIFAR-10 dataset, we utilize the convolutional layers of ResNet-20 \cite{he2016deep} as our feature extractor \cite{wilson2016stochastic} and achieve a remarkable accuracy of 95.56 on the test set. Additionally, we evaluate our approach using two large-scale classification datasets, the Higgs datase and the SUSY dataset, which are presented in Table \ref{tab:large_class}.
\begin{table}[t]
	\centering
	\caption{Mean test accuracy ($\%$) and training details achieved by DSVI, SGHMC, IPVI and DDVI (ours) DGP model for three image classification datasets. Results are shown for 3 and 4 layers as indicated, and runtime is given per iteration.}
	\resizebox{\linewidth}{!}{ 
		\centering
		\begin{tabular}{llcccccc}   
			\toprule
			Data Set & Model & Time$3$ & Iter$3$ & Acc$3$ & Time4 & Iter4 & Acc4 \\
			\midrule
			& DSVI & 0.34s & 20K & 97.17 & 0.54s & 20K & 97.41 \\
			MNIST & IPVI & 0.49s & 20K & 97.58 & 0.62s & 20K & 97.80 \\
			& SGHMC & 1.14s & 20K & 97.25 &  1.22s & 20K & 97.55 \\
			
			& DDVI & 0.38s & 20K & \textbf{98.84} & 0.50s & 20K & \textbf{99.01} \\
			\hline
			
			& DSVI & 0.34s & 20K & 87.45 & 0.50s & 20K & 87.99 \\
			Fashion & IPVI & 0.48s & 20K & 88.23 & 0.61s & 20K & 88.90 \\
			& SGHMC & 1.21s & 20K & 86.88 & 1.25s & 20K & 87.08 \\
			
			& DDVI  & 0.40s & 20K & \textbf{90.36} & 0.55s & 20K & \textbf{90.85} \\
			
			\hline
			
			& DSVI & 0.43s & 20K & 91.47 & 0.66s & 20K & 91.79 \\
			CIFAR-10 & IPVI & 0.62s & 20K & 92.79 & 0.78s & 20K & 93.52 \\
			& SGHMC & 8.04s & 20K & 92.62 & 8.61s & 20K & 92.94 \\
			
			& DDVI & 0.45s & 20K & \textbf{95.23} & 0.69s & 20K & \textbf{95.56} \\
			
			\bottomrule
	\end{tabular}}
	
	\label{tab:image-classification}
\end{table}
\begin{table}[th]
	\caption{Test AUC values for large-scale classification datasets. Uses random 90\% / 10\% training and test splits.}
	\centering
	\resizebox{0.8\columnwidth}{!}{
		\begin{tabular}{ l r c c }
			\toprule
			& & SUSY & HIGGS \\
			\cmidrule(lr){3-4}
			& $N$ & 5,500,000 & 11,000,000\\
			& $D$ & 18 & 28 \\
			
			\midrule
			\multirow{4}{*}{\shortstack{DSVI  \\ $M=128$}} & $L=2$      & 0.876 & 0.830 \\
			& $L=3$      & 0.877 & 0.837 \\
			& $L=4$      & 0.878 & 0.841 \\
			& $L=5$      & 0.878 & 0.846 \\
			\midrule
			\multirow{4}{*}{\shortstack{IPVI \\ $M=128$}} & $L=2$      & 0.879 & 0.843 \\
			& $L=3$      & 0.882 & 0.847 \\
			& $L=4$      & 0.883 & 0.850 \\
			& $L=5$      & 0.883 & 0.852 \\
			\midrule
			\multirow{4}{*}{\shortstack{SGHMC\\ $M=128$}} & $L=2$      & 0.879 & 0.842 \\
			& $L=3$      & 0.881 & 0.846 \\
			& $L=4$      & 0.883 & 0.850 \\
			& $L=5$      & 0.884 & 0.853 \\
			\midrule
			\multirow{4}{*}{\shortstack{DDVI \\ $M=128$}} & $L=2$      & \textbf{0.883} & \textbf{0.849} \\
			& $L=3$      & \textbf{0.885} & \textbf{0.852} \\
			& $L=4$      & \textbf{0.887} & \textbf{0.856} \\
			& $L=5$      & \textbf{0.886 }& \textbf{0.857} \\
			\midrule
		\end{tabular}
	}
	\label{tab:large_class}
\end{table}

\begin{table}
	\centering
	\caption{Mean RMSE and NLL  achieved by DSVI, SGHMC, IPVI and DDVI (ours) GPLVM model for data recovery task. Standard deviation is shown in parentheses. Runtime is given per iteration.}
	\small
	\resizebox{\linewidth}{!}{ 
		\centering
		\begin{tabular}{llcccc}   
			\toprule
			Data Set & Model & Time  & Iter & RMSE  & NLL  \\
			\midrule
			& DSVI & 0.32s & 20K & 8.32 (0.2) & 1.49 (0.02)  \\
			Frey Faces & IPVI & 0.42s & 20K & 7.91 (0.4) & 1.33 (0.02)  \\
			& SGHMC & 1.13s & 20K & 7.95 (0.3)  & 1.36 (0.03)    \\
			& DDVI & 0.36s & 20K & \textbf{7.64} (0.2) & \textbf{1.17} (0.01)   \\
			\bottomrule
	\end{tabular}}
	\label{tab:comparison2}
	
\end{table}
\begin{figure}
	\begin{center}
		\includegraphics[width=7.5cm]{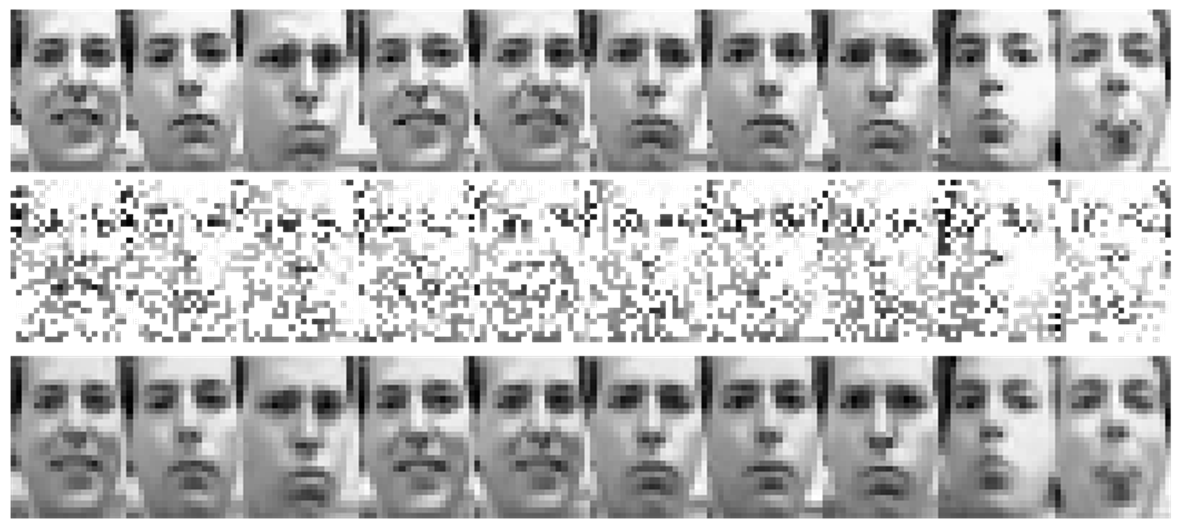}
		\caption{The Brendan faces reconstruction task with 75\% missing pixels. The top row represents the ground truth data and the bottom row showcases the reconstructions from the 20-dimensional latent distribution.}
		\label{fig3}
	\end{center}
\end{figure}

\subsection{Unsupervised Learning for Data Recovery Task}
We conducted a reconstruction experiment on  Frey Faces Data \cite{roweis2000nonlinear}, focusing on how models capture uncertainty when training with missing data in structured inputs. We used the entire dataset  with a latent variable dimensionality of 20. The image data set contains 1965 images of a face taken from sequential frames of a short video. Each image is of size 20$\times$28
yielding a 560 dimensional data space. In both cases, we chose 5\% of the training set as missing data samples and removed 75\% of their pixels, seeking to recover their original appearance. Figure \ref{fig3} summarize the samples generated from the learned latent distribution. This reconstruction experiment is performed using the Gaussian Process Latent Variable Model (GPLVM) \cite{titsias2010bayesian} and is similar to the related work by \cite{gal2014distributed}.  

To demonstrate the effectiveness of our method in producing more accurate likelihoods on image datasets, we present in Table \ref{tab:comparison2} negative log-likelihood, and RMSE for reconstructed images on the Frey Faces, comparing with baseline methods. The results show that our method  converges to higher likelihoods and lower RMSE, indicating superior performance in high-dimensional and multi-modal image data. This suggests that adding DDVI method can also improve the convergence of the traditional GPLVM methods.

\section{Conclusion }
We have introduced Denoising Diffusion Variational Inference (DDVI) as an alternative approach for inferring the posterior distribution of inducing points in Deep Gaussian Processes (DGPs). By employing a denoising diffusion stochastic differential equation (SDE) and utilizing the score matching method, we are able to accurately approximate challenging score functions using a neural network. Through extensive experiments and comparisons with baseline methods on various datasets, we demonstrated the effectiveness of DDVI in posterior inference of inducing points for DGP models. The DDVI method addressed the limitations of traditional variational inference techniques, reducing biases and improving accuracy in the posterior approximation. Our proposed DDVI approach not only enhances computational efficiency, but also provides a more robust framework for Bayesian deep learning with DGPs.

\section*{Acknowledgements}

The work is supported by the Fundamental Research Program of Guangdong, China, under Grant 2023A1515011281; and in part by the National Natural Science Foundation of China under Grant 61571005.
\section*{Impact Statement}
This paper aims to contribute to the advancement of the Machine Learning field. Our work may have various potential societal implications, none of which we believe need to be specifically emphasized here.

\newpage
\appendix
\section{Additional Related Works}
A line of closely related work focusing on enhancing variational posteriors with decoupled/orthogonal inducing points has garnered significant attention in recent years \cite{cheng2017variational, salimbeni2018orthogonally, shi2020sparse, sun2021scalable}. These methods present variational Gaussian process models that segregate the representation of mean and covariance functions within the reproducing kernel Hilbert space. This novel parametrization extends previous models and allows for solving the variational inference problem using stochastic gradient ascent with linear time and space complexity in the number of mean function parameters. In contrast to these approaches, our work diverges in its emphasis on precise posterior inference for Gaussian process models rather than complexity analysis concerning inducing points.

Another line of related work is on the idea of fully Bayesian Gaussian processes \cite{lalchand2020approximate,rossi2021sparse}, particularly \cite{rossi2021sparse} have applied MCMC-related methods and the ideas of fully Bayesian approaches to deep GPs, achieving significant improvements. Our work, on the other hand, focuses on improvements and advancements in the field of variational inference.

Additionally, another line of related work is on implict stochastic processes \cite{ma2019variational}, that is not restricted to Gaussian predictive distributions, which is closely related to function-space models \cite{sun2019functional, mescheder2017adversarial}. Furthermore, \cite{ortega2022deep} further extends the idea of DGP models to implict stochastic processes, enhancing the flexibility of the models. We believe that this is a valuable complement to our approach, and future work may involve integrating our method with these advancements to explore more useful applications.
 
\end{document}